# Adaptive Learning of Design Strategies over Non-Hierarchical Multi-Fidelity Models via Policy Alignment

**Akash Agrawal**

Department of Mechanical Engineering,
Carnegie Mellon University,
5000 Forbes Avenue,
Pittsburgh, PA 15213
e-mail: aragrawa@andrew.cmu.edu

**Christopher McComb[1]**

Department of Mechanical Engineering,
Carnegie Mellon University,
5000 Forbes Avenue,
Pittsburgh, PA 15213
e-mail: ccm@cmu.edu

## ABSTRACT

*Multi-fidelity Reinforcement Learning (RL) frameworks significantly enhance the efficiency of engineering design by leveraging analysis models with varying levels of accuracy and computational costs. The prevailing methodologies, characterized by transfer learning, human-inspired strategies, control variate techniques, and adaptive sampling, predominantly depend on a structured hierarchy of models. However, this reliance on a model hierarchy overlooks the heterogeneous error distributions of models across the design space, extending beyond mere fidelity levels. This work proposes ALPHA (Adaptively Learned Policy with Heterogeneous Analyses), a novel multi-fidelity RL framework to efficiently learn a high-fidelity policy by adaptively leveraging an arbitrary set of non-hierarchical, heterogeneous, low-fidelity models alongside a high-fidelity model. Specifically, low-fidelity policies and their experience data are dynamically used for efficient targeted learning, guided by their alignment with the high-fidelity policy. The effectiveness of ALPHA is demonstrated in analytical test optimization and octocopter design problems, utilizing two low-fidelity models alongside a high-fidelity one. The results highlight ALPHA's adaptive capability to dynamically utilize models across time and design space, eliminating the need for scheduling models as required in a hierarchical framework. Furthermore, the adaptive agents find more direct paths to high-performance solutions, showing superior convergence behavior compared to hierarchical agents.*

---

[1] Corresponding author

1. **INTRODUCTION**

Multi-fidelity Reinforcement Learning (RL) frameworks [1–5] enhance design efficiency by leveraging high-fidelity simulations and low-fidelity alternatives like simplified physics simulations, reduced-order models, surrogate models, and partially converged results [6–10]. These frameworks, which employ strategies such as transfer learning [1,2], human-inspired approaches [3], control variate techniques [4], and adaptive sampling [5], rely on a hierarchical structure of models. This structure entails discrete levels of model fidelity, from low to high, with each level representing a distinct trade-off between computational cost and accuracy. However, this approach often overlooks the nuanced error distributions across the design space that transcend these discrete fidelity levels. These error distributions stem from diverse sources inherent to different types of low-fidelity models. They can vary widely due to factors such as model-specific assumptions and simplifications [6–9], discrepancies in model complexity versus the complexity of different regions within the design space [11–13], the variability in data availability and distribution across models [14–17] and challenges posed by hybrid modeling approaches [18,19]. To address this challenge, this work proposes an adaptive multi-fidelity RL framework, ALPHA (Adaptively Learned Policy with Heterogeneous Analyses), that leverages heterogeneous low-fidelity models alongside a high-fidelity model to learn a high-fidelity policy.

Deep RL [20–23] is adept at exploring complex, non-differentiable, multi-dimensional designs spaces involving various objectives and constraints [3,24–33]. It excels in sequential decision-making, incrementally navigating through complex spaces for optimal long-term outcomes. This navigation becomes more efficient when the design task is part of a higher-level iterative design procedure [34]. Moreover, the adaptivity and noise tolerance of Deep RL make it a robust choice for dynamic design environments [35,36]. While RL offers numerous possibilities for advancing engineering design, its practical implementation is limited by the computational resources required for high-fidelity simulations (like computational fluid dynamics or finite element analysis). These simulations, indispensable for the evaluation of design alternatives, serve as the reward function in RL algorithms, often causing computational bottlenecks. This can delay project timelines and raise concerns over economic and environmental sustainability [37,38], limiting their applicability in engineering design.

In the pursuit of advancing computational efficiency of RL in design, multi-fidelity RL is emerging as a promising approach [1–5]. This approach strategically integrates both high-fidelity simulations, which provide accurate analysis of engineered systems at higher computational costs, and low-fidelity models, which offer computationally economical but less precise alternatives [6–10]. By leveraging the strengths of both high and low-fidelity models, multi-fidelity RL facilitates more efficient design exploration, especially when high-fidelity data is limited or expensive to obtain. The state-of-the-art methods including transfer learning [1,2], human-inspired approaches [3], control variate approaches [4], and adaptive sampling [5] have shown significant promise in diminishing computational demands. These methods leverage prior knowledge and/or implicit information acquired during the learning process regarding model accuracy. This insight, combined with key aspects of the multi-fidelity RL methodology, enables strategic decisions on model usage, balancing computational cost and solution quality of designs. However, these approaches predominantly rely on a rigid hierarchical structure of models. This hierarchical view assumes a monotonic relationship between model fidelity and accuracy, with each step up in fidelity offering superior performance across the entire design space. While this hierarchy simplifies model management, it oversimplifies the nuanced differences that can exist between models. In this work, we examine these models from the perspective of the error of the model over the design space. These errors are not uniformly distributed; instead, they can exhibit heterogeneity, varying significantly across different regions of the design space [10,39,40].

The heterogeneity of errors in low-fidelity models stems from diverse sources inherent to different types of low-fidelity models. For simplified physics models, the underlying assumptions may ignore or oversimplify certain phenomena that become critical in some regions of the design space [11–13]. For instance, the geometry of a design significantly influences the dominance of laminar or turbulent flow. This variation in flow type across different regions affects model accuracy because different models are typically calibrated for specific flow conditions. A model that performs well under laminar flow might not accurately capture the complexities of turbulent flow. Therefore, using a model that does not account for the dominant flow type in a particular region of the design space can lead to errors. Moreover, different low-fidelity models might adopt varying assumptions about the flow type and employ distinct methods to model that flow type, impacting model accuracy [11–13].

Surrogate model frameworks for multi-fidelity RL alleviate some of the issues noted above, but are severely limited by the availability and distribution of data [14–17], especially in engineering design applications. The surrogate models available for exploration may have been prepared using distinct datasets, some of which may be more densely populated in certain regions of the design space. This disparity in the underlying datasets across the design space can lead to diverse error distributions of models. Furthermore, datasets with partially converged results of high-fidelity simulations, may also lead to heterogeneity in model errors. Hybrid models, which combine different modeling approaches, are typically used to reduce errors by leveraging the complementary strengths of each approach. For example, integrating empirical data with physics-based simulations [19] can enhance model accuracy by combining data-driven insights with the robustness of physical laws. However, when the complementarity between these approaches is not perfect, the hybrid model may still exhibit varying levels of accuracy across different regions of the design space. In such cases, the model might introduce additional complexity, as the different components of the hybrid model may have their own unique error characteristics that can interact in unpredictable ways [18]. To address such challenges, this work introduces an adaptive multi-fidelity RL framework, ALPHA (Adaptively Learned Policy with Heterogeneous Analyses), that leverages heterogeneous low-fidelity models alongside a high-fidelity model to learn a high-fidelity policy.

The rest of the paper is organized as follows. Section 2 introduces multi-fidelity RL and discusses several prevailing frameworks. In Section 3, we propose an adaptive multi-fidelity RL framework (ALPHA) and detail other methodologies used in this work. In Section 4, an analytical test optimization problem and an octocopter design problem are introduced to demonstrate the effectiveness of the proposed framework. Section 5 presents the results of the case studies, including the patterns of model usage across time and design space, the evolution of policies across training and a comparison with a hierarchical framework. Section 6 summarizes the contribution of the paper and proposes several directions for future work.

## 2. BACKGROUND

Multi-fidelity reinforcement learning is an approach that leverages data from multiple models of varying fidelity to enhance computational efficiency, particularly in contexts where high-fidelity data is scarce or costly to obtain. The term fidelity [41] denotes the accuracy or realism of a model, where high-fidelity models provide more accurate or realistic data at the expense of higher computational costs or time consumption. On the other hand, low-fidelity models, though less accurate, are computationally cheaper to interact with, offering a simplified representation of reality.

In the realm of engineering design, simulators are often used for training RL agents. Here, the notion of simulator fidelity is pivotal in characterizing the accuracy of the relationship between an engineered system and its performance attributes. High-fidelity simulators, often employing computationally intensive numerical methods (like computational fluid dynamics or finite element analysis), accurately capture the underlying relationships of interest, mirroring the behavior of real-world systems closely.

Conversely, low-fidelity simulators, which could entail simplified physics or partially converged results, offer a less accurate but computationally economical alternative. Low-fidelity models, such as reduced-order models and surrogate models, also provide a computationally efficient means to approximate the behavior of more complex systems with less accuracy [6–10].

Multi-fidelity model-based and model-free reinforcement learning techniques have emerged as paradigms to harness varying fidelity levels in several fields including engineering design, robotics, and optimization. While all techniques rely on being aware of the highest fidelity level, several utilize additional insights about the hierarchy of simulators, including bounds on reward errors across fidelity levels [42]. Several methods leverage transfer learning approaches which can be further classified into unidirectional and bidirectional transfer. The former uses some heuristic to control the switch to higher levels, however it is typically not informed by knowledge across fidelity levels. For instance, Bhola et al. [1] utilized a reward convergence metric as the control criteria to switch to a higher fidelity level in an airfoil design problem tackled with Proximal Policy Optimization (PPO). In another work by Geng et al. [2], both the design configuration and policies were transferred from low to high fidelity in a Deep Deterministic Policy Gradient solver for a propeller design problem.

On the other hand, bidirectional transfer involves more explicit learning across fidelities by intricately tying fidelity switching to the learning of the transition and reward models across the levels in model-based approaches. For instance, Cutler et al. [42] employed a greedy tabular $Q$ learning approach, wherein the transfer of $Q$ values from

low- to high-fidelity happens once a predefined number of steps exhibiting sufficient model learning of the former is encountered. Furthermore, while operating at a higher-fidelity level, the algorithm also informs updates to the models at the lower level. Concurrently, at each step, the algorithm checks the certainty of the models at this level. If uncertainty is encountered, it steps down to the lower level to enable continual learning including that of previously unnoticed dynamics. While using a greedy algorithm, the overall operation across fidelity levels still ensures a degree of exploration. Another work [43] extends this by employing Gaussian Processes (GPs) to learn the transition function, whereby the underlying variance, and the sum of variances of past experiences, informs the fidelity switching decisions. Moreover, they also investigated a model-free variant that learns optimal $Q$ values using GPs.

Diverging from transfer-based methods, Felipe Leno da Silva et al. [44] employed elitist heuristics in symbolic optimization problems, wherein better performers are biased to operate at high fidelity levels. Further, they utilize modified policy gradient objectives that are biased towards high-fidelity returns or conditioned with respect to a top-quantile $Q$ measure. This technique showcases superior performance than transfer-based approaches. In another work, a multi-fidelity action value function estimator that is unbiased with respect to the highest fidelity is designed [4]. Specifically, it is based on a control variate approach that utilizes additional low-fidelity experience for the true low-fidelity estimate. It taps into the correlation between low and high-fidelity returns, leading to a reduced variance in $Q$ estimates and a better resultant policy. In another line of work by Qiu et al. [45], a multi-agent study employs a depth-first search strategy to unearth local feasible policies on a low-fidelity simulator, which is further leveraged to formulate mixed policies alongside a baseline soft actor-critic policy for enhanced policy rollout.

In contrast to previous approaches that independently leveraged models of varying fidelity, Li et al. [5] utilized a low-fidelity and a multi-fidelity surrogate for training a DQN agent on a design problem. Initially, the agent learns with a low-fidelity model that is pretrained with low-fidelity data. This provides a broad understanding of the design space. Further, this trained agent guides the selective sampling of high-fidelity data, focusing on the most promising regions identified during the initial exploration. The combined low- and high-fidelity data are subsequently used to train a multi-fidelity surrogate. This multi-fidelity surrogate is employed to continue the training of the agent, ultimately leading to high-quality solutions at a reduced computational expense.

This work specifically builds on an earlier hierarchical framework [3] that progressively utilizes models from low- to high-fidelity within episodic design tasks to achieve a high solution quality at a reduced computational cost. In the current work, we extend the framework to adaptively handle heterogeneous low-fidelity models alongside a high-fidelity one to learn a high-fidelity policy.

3. METHODOLOGY

This work proposes a multi-fidelity RL framework, ALPHA (Adaptively Learned Policy with Heterogeneous Analyses), for design space exploration that can adaptively handle heterogeneous low-fidelity models alongside a high-fidelity. This section outlines the methodology used to construct and evaluate the framework. In Section 3.1, we propose the adaptive multi-fidelity framework based on the alignment of low-fidelity policies with the high-fidelity policy. Section 3.2 details the methodology for training and evaluating RL agents to assess their ability to adaptively utilize models across time and design space. Further, we describe a comparative study that evaluates the performance of ALPHA against a hierarchical framework.

**3.1 ALPHA, An Adaptive Multi-fidelity Reinforcement Learning Framework**

The adaptive multi-fidelity RL framework proposed in this work, referred to as ALPHA (Adaptively Learned Policy with Heterogeneous Analyses), builds upon a prior multi-fidelity framework [3]. Like that work, ALPHA aims to solve a skeletal design problem by starting with seed designs and iteratively tuning the continuous and discrete variables to minimize an objective, $f'$ while satisfying the constraints $\boldsymbol{G}'$ and $\boldsymbol{H}'$. Furthermore, similar to the previous framework and the techniques discussed in Section 2, we assume awareness of the highest fidelity level. However, unlike the previous framework in which a single agent controlled the entire process, ALPHA uses one agent for every analysis model, catering to the heterogeneity of models rather than using a hierarchical structure. Each design agent learns to take actions ($\boldsymbol{a}_t$) in a design state ($\boldsymbol{s}_t$) based on the feedback received in the form of scalar rewards (r). Specifically, each agent learns a policy ($\pi$) to maximize the sum of rewards derived from the corresponding analysis model. The agent reward ($r_{t+1}$) measures the quality of the action $\boldsymbol{a}_t$ that transitions the design from state $\boldsymbol{s}_t$ to $\boldsymbol{s}_{t+1}$, with objective and constraint components like the prior work [3]. The adaptive interaction of the agents with the design space is illustrated in Figure 1 and detailed in Algorithm 1.

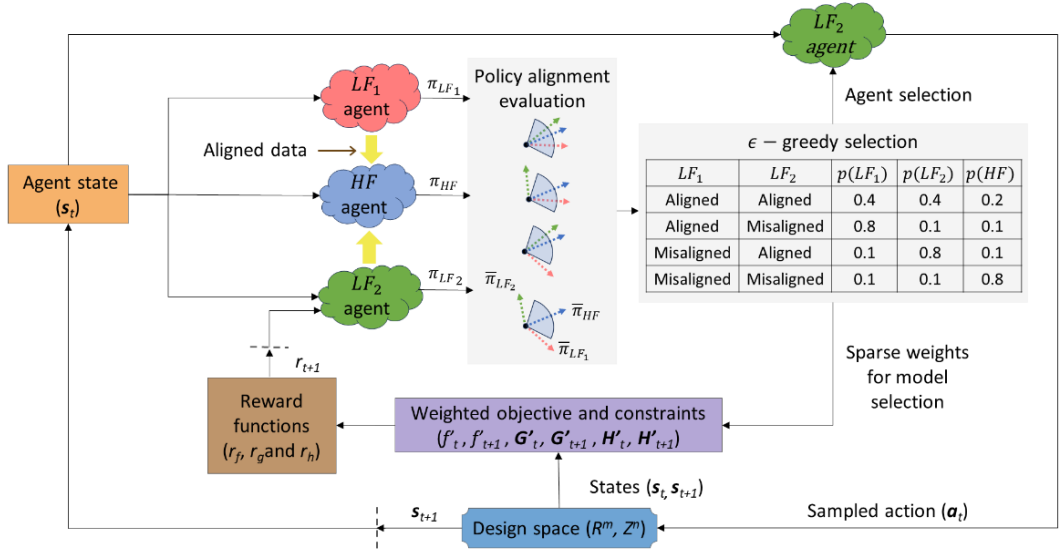

FIGURE 1: ALPHA (ADAPTIVELY LEARNED POLICY WITH HETEROGENEOUS ANALYSES)

ALPHA comprises one high-fidelity ($HF$) agent and multiple low-fidelity ($LF_i$) agents, with the case of two LF agents ($LF_1$ and $LF_2$) illustrated in Figure 1 and discussed in detail here. These agents interact with the design environment, collecting experience data used to update their respective policies. This experience data contains states, actions, and rewards, observed during interactions with the design space. During training, the degree of alignment between different $LF$ policies and the $HF$ policy will vary based on the underlying heterogeneity of the error distributions of the $LF$ models. The proposed approach aims to prioritize using the $HF$ agent in regions where all $LF$ policies diverge significantly from the HF policy, ensuring targeted learning in critical areas.

Conversely, regions where one or more of the $LF$ policies align well with the $HF$ policy can continue to learn using the aligned $LF$ agents, thereby reducing computational costs.

For the case of two low-fidelity models, the alignment of both the low-fidelity policies $\pi_{LF_1}$ and $\pi_{LF_2}$ are measured with respect to the high-fidelity policy $\pi_{HF}$. Specifically, cosine similarity between the mean of the action distributions of the policies ($\bar{\pi}_{LF_1}, \bar{\pi}_{LF_2}, \bar{\pi}_{HF}$) serves as the alignment metric. Further, an alignment threshold determines whether a particular $LF$ model is aligned to the $HF$ model or not. For instance, for the iteration illustrated in Figure 1, the alignment threshold with respect to the $HF$ policy corresponds to an angle of 45 degrees. For this case of two low-fidelity models, there are four possible scenarios based on the alignment threshold: both $LF_1$ and $LF_2$ align with $HF$; $LF_1$ aligns, but $LF_2$ does not; $LF_2$ aligns, but $LF_1$ does not; neither $LF_1$ nor $LF_2$ align with $HF$.

***Algorithm 1***: ***ALPHA*** (***Adaptively Learned Policy with Heterogeneous Analyses***)
---
$\pi_{HF}, \pi_{LF_1}, \pi_{LF_2} \leftarrow \pi_0$ (policy functions); $v_{HF}, v_{LF_1}, v_{LF_2} \leftarrow v_0$ (value functions)
$B_{HF}, B_{LF_1}, B_{LF_2} \leftarrow [\,]$ (data buffers), $\epsilon \coloneqq constant$ (Model choice $\epsilon$-greedy parameter)

**for** $e \subset \{0, 1, ...., EPISODE\ COUNT - 1\}$ **do**
    $s_0 \leftarrow$ sample seed design
    **for** $t \subset \{0, 1, ...., EPISODE\ LENGTH - 1\}$ **do**
        $S_{\cos,1} \leftarrow$ cosine similarity between $\bar{\pi}_{LF_1}(s_t)$ and $\bar{\pi}_{HF}(s_t)$
        $S_{\cos,2} \leftarrow$ cosine similarity between $\bar{\pi}_{LF_2}(s_t)$ and $\bar{\pi}_{HF}(s_t)$
        $model_t, aligned_t \leftarrow$ **model choice** ($e, S_{cos,1}, S_{cos,2}, \epsilon$)      (see Algorithm 2)
        $a_t \leftarrow$ sample action from $\pi_{model_t}(s_t)$
        $s_{t+1}, r_{t+1} \leftarrow step_{model_t}(s_t, a_t)$
        **if** $t > 0$ AND $model_t = model_{t-1}$
            $B_{model_t} \ll (s_t, a_t, r_{t+1}, s_{t+1}, aligned_t)$      (append to current sequence)
        **else**
            $B_{model_t} \mathrel{+}= [(s_t, a_t, r_{t+1}, s_{t+1}, aligned_t)]$      (start new sequence)
        $s_t \leftarrow s_{t+1}$
    **end**

    Augment $HF$ data with $LF$ data where there is alignment
    **for** $model$ in $LF_1, LF_2$ **do**
        **for** $sequence$ in $B_{model}$ **do**
            $subsequences \leftarrow$ find contiguous subsets in $sequence$ where $aligned = True$
            $B_{HF} \mathrel{+}= [subsequence$ **for** $subsequence$ in $subsequences]$
        **end**
    **end**

    **for** $model$ in $LF_1, LF_2, HF$ **do**
        **if** $BATCH$ data collected in $B_{model}$
            train $\pi_{model}$ and $v_{model}$ using $BATCH$ and policy optimization algorithm (e.g. $PPO$)
            $B_{model} \leftarrow [\,]$
    **end**
**end**

Based on this, model choice is managed using an $\epsilon$-greedy strategy (Algorithm 2), which balances exploration of different models with the exploitation of their current knowledge. This ensures that no region in the design space gets fixated to use a specific model with limited overall experience of the agents. After selecting a specific model fidelity ($LF_2$ in Figure 1), an action is sampled from the chosen policy to take a step in the design space. Further, for the selected fidelity, the objectives and constraints are evaluated to get the resultant reward function, thereby completing the overall learning timestep. Lastly, the alignment threshold follows a cosine schedule [46] on the underlying angle to avoid a drastic decrease of the reliance on $LF$ models in the beginning, while allowing for more refined solutions using the $HF$ model towards the end (adjusting from 90 to 0 degrees).

---
**Algorithm 2**: *Model choice*

**function** model choice ($e, S_{cos,1}, S_{cos,2}, \epsilon$)

$$\theta \leftarrow \begin{cases} \cos\left[\frac{\pi}{4}\left(1 + \cos\left(\pi \cdot \frac{e}{0.9 \times EP_{MAX}}\right)\right)\right], & e < 0.9 \times EP_{MAX} \\ 0, & e \geq 0.9 \times EP_{MAX} \end{cases}$$

$$(p_{LF_1}, p_{LF_2}, p_{HF}) \leftarrow \begin{cases} \left(\frac{1-\epsilon}{2}, \frac{1-\epsilon}{2}, \epsilon\right), & S_{cos,1} > \theta \text{ AND } S_{cos,2} > \theta \\ \left(1-\epsilon, \frac{\epsilon}{2}, \frac{\epsilon}{2}\right), & S_{cos,1} > \theta \text{ AND } S_{cos,2} < \theta \\ \left(\frac{\epsilon}{2}, 1-\epsilon, \frac{\epsilon}{2}\right), & S_{cos,1} < \theta \text{ AND } S_{cos,2} > \theta \\ \left(\frac{\epsilon}{2}, \frac{\epsilon}{2}, 1-\epsilon\right), & S_{cos,1} < \theta \text{ AND } S_{cos,2} < \theta \end{cases}$$

$$model \leftarrow Categorical\bigl(\{LF_1, LF_2, HF\}, \{p_{LF_1}, p_{LF_2}, p_{HF}\}\bigr)$$

$$aligned = \left[model == (\underset{m \subset \{LF_1, LF_2, HF\}}{\operatorname{argmax}} p_m)\right]$$

**return** *model, aligned*

---

To achieve a unified high-fidelity policy capable of comprehensively exploring the design space, the experience data from $LF$ policies is integrated into the learning process of the $HF$ policy. Specifically, this augmentation occurs in regions where $LF$ policies are aligned with the $HF$ policy and are chosen exploitatively, ensuring that the incorporated data maintains the accuracy of the $HF$ policy. Essentially, rewards from $LF$ models are used in aligned regions of a trajectory, and $HF$ value estimates are used for beyond, maintaining the learning precision of the $HF$ policy. By incorporating $LF$ experience data in the aligned regions, the $HF$ agent can efficiently learn a unified high-fidelity policy capable of comprehensively exploring the design space.

**3.2 Training and Evaluating RL Agents**

A proximal policy optimization algorithm [47] is used for training the agents using ALPHA with a high-fidelity model and the heterogeneous low-fidelity models. Several randomly sampled designs are used as seed designs for training. After training, the learned high-fidelity policy is evaluated by passing the seeds to yield the solutions, the quality of which is measured using the high-fidelity model. To better contextualize the results of ALPHA, we trained and evaluated policies with a hierarchical multi-fidelity RL framework from prior work [3], and with each of the low-fidelity 1 ($LF_1$), low-fidelity 2 ($LF_2$) and high-fidelity ($HF$) models individually.[3] For the hierarchical framework, two configurations were used for training: 35% steps with $LF_1$, 35% steps with $LF_2$, followed by 30% $HF$, and 35% steps with $LF_2$, 35% steps with $LF_1$, followed by 30% $HF$. These configurations were chosen to investigate how the sequence of the non-hierarchical $LF$ models affects the learning process, assuming that $HF$ is of higher fidelity than both $LF$ models across the search space. We aimed to allocate equal training time to each fidelity level, but an exact equal split was infeasible due to the total number of steps in the episode being 20. Therefore, we adjusted the allocation to 35% of the steps for $LF_1$ (7 steps), 35% for $LF_2$ (7 steps), and 30% for $HF$ (6 steps). These agents are named ALPHA, Hierarchical Multi-Fidelity RL (MFRL) 1, Hierarchical MFRL 2, High-fidelity RL, Low-fidelity 1 RL, and Low-fidelity 2 RL, and are referred to by these names in the rest of the paper. The seed designs, number of iterations per episode, the total number of episodes and other RL hyperparameters are kept the same across all the agents. Furthermore, we also visualize representative evaluation trajectories for all the agents [48] to interpret the learned policy.

In order to assess the adaptive capability of ALPHA, the model usage is tracked during training. Moreover, the policies are also saved at regular intervals during training to investigate their evolution with the aim of interpreting the underlying learning process of the agents. After completion of training, we analyze the patterns of model usage both across time and design space along with the evolution of the learned policies.

For spatial analysis, the design space is discretized into a grid and model usage proportions are computed for each cell. Moran's I [49] is a measure of spatial autocorrelation which we use here to assess whether model usage exhibits spatial clustering. Spatial clusters in model usage indicate regions of the design space where certain models are preferentially used, reflecting targeted learning. Spatial relationships between grid cells are captured using kernel distance weighting, and significance is assessed through permutation testing, which evaluates the observed Moran's I against a null distribution generated by randomizing the data across grid cells.

Lastly, we evaluate the quality-efficiency tradeoff of ALPHA in comparison with the other agents. As the number of iterations per episode and total number of episodes is constant across the agents, the total time required to evaluate the objectives and constraints using the low- and high-fidelity models is utilized to reflect the computational efficiency.

## 4. CASE STUDIES

In order to rigorously evaluate ALPHA, we define two complementary case studies. First, we consider an analytical test optimization problem, which enables us to assess the framework in a narrow and tightly-controlled environment. Second, we assess the framework on an octocopter design problem to demonstrate the framework's practical application. Both case studies utilize two low-fidelity models and one high-fidelity model. The details of the analytical test optimization problem and the octocopter design problem are described in Secs. 4.1 and 4.2, respectively.

**4.1 Analytical test optimization problem**

The analytical test optimization problem in this work is adapted from the two-dimensional Ackley function [50], a standard benchmark in optimization. Specifically, we introduce two parameters corresponding to the shifted position of the global minimum and a scaling factor for the minimum function value at this point. These modifications offer the flexibility to prepare high- and low-fidelity models as discussed further below. The modified Ackley function ($f$) is mathematically defined as follows:

$$f_{x_c,\alpha}(x_1, x_2) = g_{x_c,\alpha}(x_1, x_2) + h_{x_c}(x_1, x_2) \tag{1}$$

$$g_{x_c,\alpha}(x_1, x_2) = 20 - 20\alpha\, e^{\left\{-0.2\sqrt{0.5\left((x_1-x_{c_1})^2+(x_2-x_{c_2})^2\right)}\right\}} \tag{2}$$

$$h_{x_c}(x_1, x_2) = e - e^{\left\{0.5\left(\cos 2\pi(x_1-x_{c_1})+\cos 2\pi(x_2-x_{c_2})\right)\right\}} \tag{3}$$

where, $x_1$ and $x_2$ belong to the two-dimensional real plane ($\mathbb{R}^2$) with bounds $-d \leq x_i \leq d$, for $i = 1, 2$, $d = 32.768$, $x_c$ is the shifted location of the minima and $\alpha$ is the scaling factor for the minimum value.

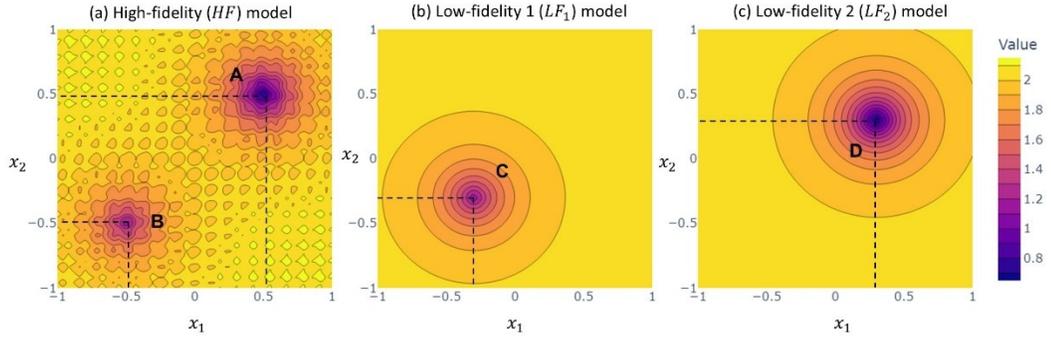

FIGURE 2: ACKLEY-BASED PROBLEM FORMULATION

We utilize the above formulation to further prepare a high-fidelity ($HF$) model and two low-fidelity ($LF_1, LF_2$) models. The contour plots of these models in the scaled domain ($-1 \leq x_i \leq 1$ for $i = 1, 2$) are shown in Figure 2. The high-fidelity model ($f_{HF}$) is prepared by the superposition of two Ackley functions, each having distinct locations and scaling factors:

$$f_{HF} = f_{(-0.5d,-0.5d),1} + f_{(0.5d,0.5d),1.5} \tag{4}$$

The locations of the global minima of the individual Ackley functions are labelled in Figure 2(a) as A and B. The mean cost of this model on an Intel Xeon CPU used in this work is 275 $\mu s$. The low-fidelity models are prepared by varying the parameters $x_c$ and $\alpha$ in $g_{x_c,\alpha}$, while only utilizing the midrange value for $h_{x_c}$. Specifically, by setting $\alpha$ as 0 for different $g_{x_c,\alpha}$ components in the low-fidelity models, each model is tailored for different regions of the search space, without following a strict fidelity hierarchy. Furthermore, the locations of the minima are shifted by a different amount than the high-fidelity model to reflect the inaccuracies that are typical of low-fidelity models. These locations are labelled as C and D in Figures 2(b, c). Lastly, simplifying $h_{x_c}$ to a midrange value reduces the computational expense while still providing a reasonable estimate on average. The two low-fidelity models are mathematically defined as follows:

$$f_{LF_1} = g_{(-0.3d,-0.3d),1} + h_m + g_{(0.3d,0.3d),0} + h_m \tag{5}$$

$$f_{LF_2} = g_{(-0.3d,-0.3d),0} + h_m + g_{(0.3d,0.3d),1.5} + h_m \tag{6}$$

$$h_m = 0.5 \times (e - e^{-1}) \tag{7}$$

The mean cost of these models on an Intel Xeon CPU used in this work is 32 $\mu$s.

The architecture of the policy neural network used in this case study is defined as follows:

$$I_2 - D_{1024,R} - D_{1024,R} - \begin{cases} (O_1)_{2,T} \\ (O_2)_{2,SP} \end{cases}$$

where $I_{N_i}$ represents the input layer of size $N_i$, $D_{j,k}$ represents a dense (hidden) layer of size $j$ with an activation denoted by $k$, $R$ represents the ReLU activation, $T$ represents the hyperbolic tangent activation, $SP$ denotes the soft plus activation and $O_{N_{o,k}}$ represents the output layers of size $N_o$ with activation $k$. Specifically, the branches $O_1$ and $O_2$ correspond to the mean and standard deviation of the action distributions that are output by the policy. Similarly, the architecture of the value function neural network used in this case study is defined as follows:

$$I_2 - D_{1024,R} - D_{1024,R} - O_{L,1}$$

**4.1 Octocopter design problem**

The octocopter design problem involves a corpus of components and a flight dynamics simulator [51,52]. The components include batteries, motors, and propellers. The design space of the problem comprises a continuous variable for arm length and three ordinal variables for the choice of batteries, motors, and propellers from an ordered set of the components. Figure 3 illustrates the design artifact of an octocopter generated by assigning random values to the design variables. The reader is referred to prior work [52] for details on the corpus of components used in this problem. By considering all possible discrete values and merely 10 values for the continuous variable, the size of the combinatorial space is of the order of $10^6$.

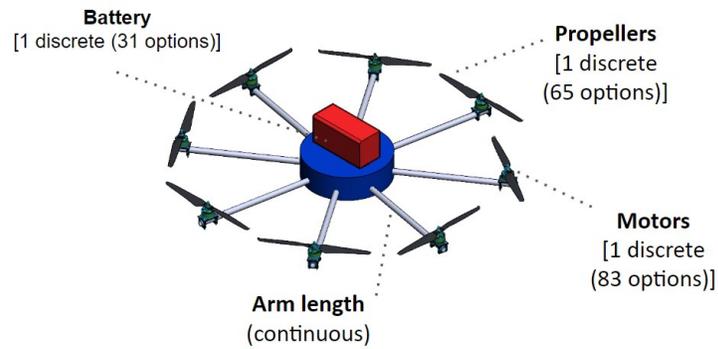

*FIGURE 3: DESIGN ARTIFACT OF OCTOCOPTER (LABELS INDICATE THE NUMBER OF DESIGN VARIABLES ASSOCIATED WITH DIFFERENT COMPONENTS)*

The design objective is based on a maneuvering task along a trajectory defined by a set of waypoints. Specifically, we define a maximization objective as follows:

$$Q = \frac{d}{D} \times \overline{v} \times (1 - \overline{e}) \tag{8}$$

where $d$ is the distance covered along the trajectory, $D$ is total distance to be covered to complete the trajectory, $\overline{v}$ is the normalized average speed of the maneuver, and $\overline{e}$ is the normalized average error (deviation) from the specified trajectory. To emphasize, this objective aims at developing long-range, fast and stable quadcopters. Figures 4-7 showcase the simulated trajectories for four exemplar octocopter configurations. Specifically, the components of each octocopter are listed. Further, the path followed by them during the simulation is visualized in three orthogonal planes along with the reference path. The reader is referred to prior work [51] on the flight dynamics simulator for further details.

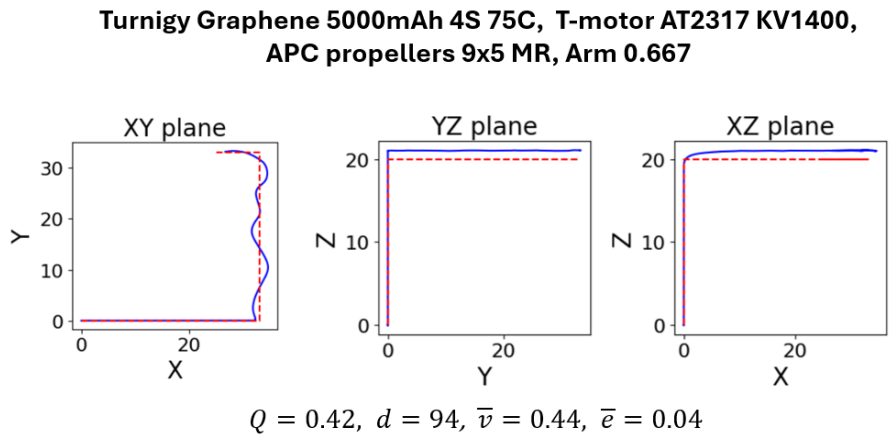

$Q = 0.42,\ d = 94,\ \overline{v} = 0.44,\ \overline{e} = 0.04$

FIGURE 4: SIMULATION OF EXEMPLAR OCTOCOPTER 1

**Turnigy Graphene 6000mAh 4S75C, T-motor AS2820 KV1050, APC propellers 13x5 5MR, Arm 0.537**

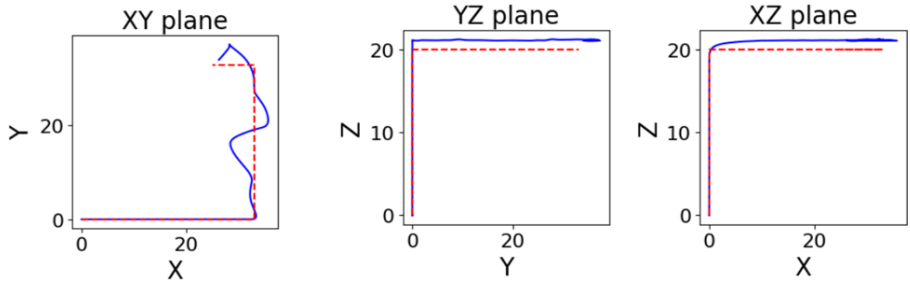

$Q = 0.37, \ d = 94, \ \overline{v} = 0.41, \ \overline{e} = 0.09$

*FIGURE 5: SIMULATION OF EXEMPLAR OCTOCOPTER 2*

**Turnigy Graphene 6000mAh 6S75C, T-motor AT2826KV1100 APC propellers 12x5 5MR, Arm 0.625**

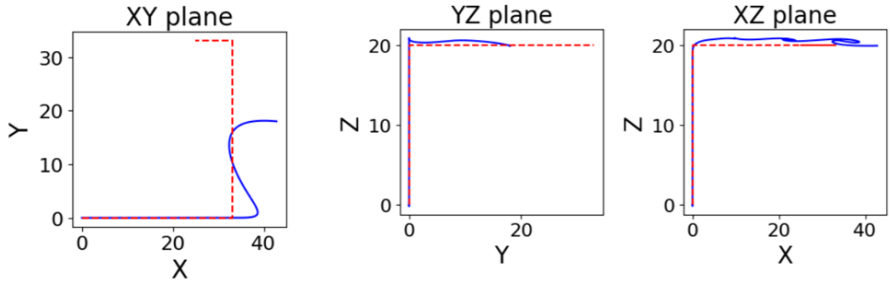

$Q = 0.09, \ d = 73, \ \overline{v} = 0.49, \ \overline{e} = 0.75$

*FIGURE 6: SIMULATION OF EXEMPLAR OCTOCOPTER 3*

**Turnigy Graphene 2200mAh 3S75C, T-motor MN5212 KV340 APC propellers 13x4 5MR, Arm 0.559**

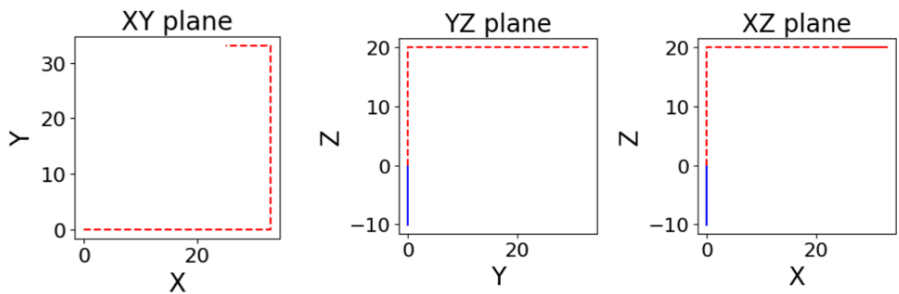

$Q = 0, \ d = 0, \ \overline{v} = 0, \ \overline{e} = 0.98$

*FIGURE 7: SIMULATION OF EXEMPLAR OCTOCOPTER 4*

The flight dynamics simulator serves as high-fidelity ($HF$) model for this case study. The median cost of this model on an Intel Xeon CPU used in this work is 1.78 s. Additionally, the cost at the 25th percentile is 0 s, while at the 75th percentile it is 22.04 s, indicating significant variability in the range and speed of the octocopters. Notably, the zero second values represent octocopters that are not flyable due to interferences or incompatibility of the components. The low-fidelity models ($LF_1, LF_2$) are prepared by training two neural networks on distinct datasets obtained from a partially converged optimization run for designing the octocopter. This optimization data was scaled using a min–max normalization technique. Figure 8 showcases the reduced dimensional Principal Component Analysis (PCA) space of this partially converged scaled data with the highlighted subsets reflecting the datasets used to train the low-fidelity models. The tailoring of each model to different regions of the search space leads to heterogenous models, contrasting with a rigid hierarchy and providing an effective testbed for the proposed framework.

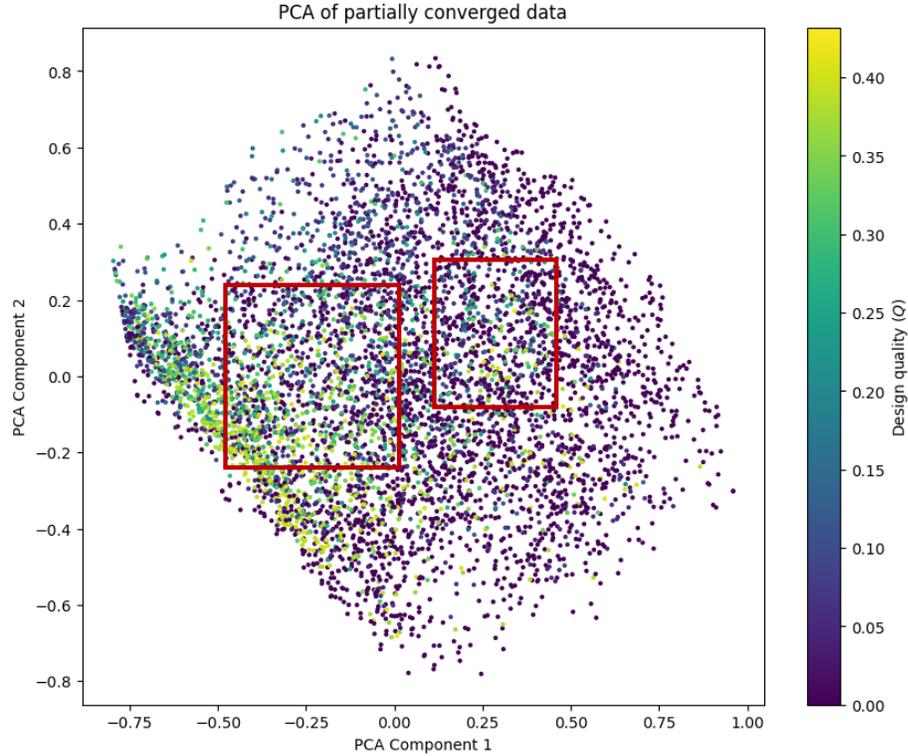

FIGURE 8: DATASETS USED TO TRAIN LOW-FIDELITY SURROGATES

The architecture of both the $LF_1$ and $LF_2$ neural networks is defined as follows:

$$I_4 - D_{64,R} - D_{32,R} - O_{1,L}$$

where $I_{N_i}$ represents the input layer of size $N_i$, $D_{j,k}$ represents a dense (hidden) layer of size $j$ with an activation denoted by $k$, $R$ represents the ReLU activation, $L$ represents linear activation, and $O_{N_o,k}$ represents the output layer of size $N_o$ with activation k. The size of the datasets used to train the $LF_1$ and $LF_2$ models are 680 and 1358 respectively. The prediction accuracies of the $LF_1$ and $LF_2$ models on the validation subsets of their respective datasets are 0.51 and 0.56, respectively. Furthermore, their prediction accuracies on the entire dataset shown in Figure 4 are 0.23 and 0.38, respectively. This implies that the low-fidelity models are indeed tailored to different regions of the search space. Lastly, the mean cost of evaluation for both these models on an Intel Xeon CPU used in this work is 208 μs.

The architecture of the policy neural network used in this case study is like the previous case and is defined as follows:

$$I_4 - D_{1024,R} - D_{1024,R} - \begin{cases} (O_1)_{4,T} \\ (O_2)_{4,SP} \end{cases}$$

Further, the architecture of the value function neural network used in this case study is like the previous case and is defined as follows:

$$I_4 - D_{1024,R} - D_{1024,R} - O_{L,1}$$

## 5. RESULTS AND DISCUSSION

### 5.1 Analytical test optimization problem

The RL policies were trained and evaluated with the proposed adaptive multi-fidelity RL framework (ALPHA), a hierarchical multi-fidelity RL framework from prior work [3], and with each of the low-fidelity 1 ($LF_1$), low-fidelity 2 ($LF_2$) and high-fidelity ($HF$) models individually. For the hierarchical framework, two configurations were used for training: 35% steps with $LF_1$, 35% steps with $LF_2$, followed by 30% $HF$, and 35% steps with $LF_2$, 35% steps with $LF_1$, followed by 30% $HF$. Specifically, 300 seed points were randomly sampled and utilized for training and evaluating all the policies. The results of the evaluation for the six cases (ALPHA, Hierarchical MFRL 1, Hierarchical MFRL 2, High-fidelity RL, Low-fidelity 1 RL, and Low-fidelity 2 RL) as per the high-fidelity model ($Q_{HF}$) are shown in Figure 9. Furthermore, Figure 10 depicts representative evaluation trajectories for all cases in the scaled problem domain, with the model contours overlaid on these trajectories.

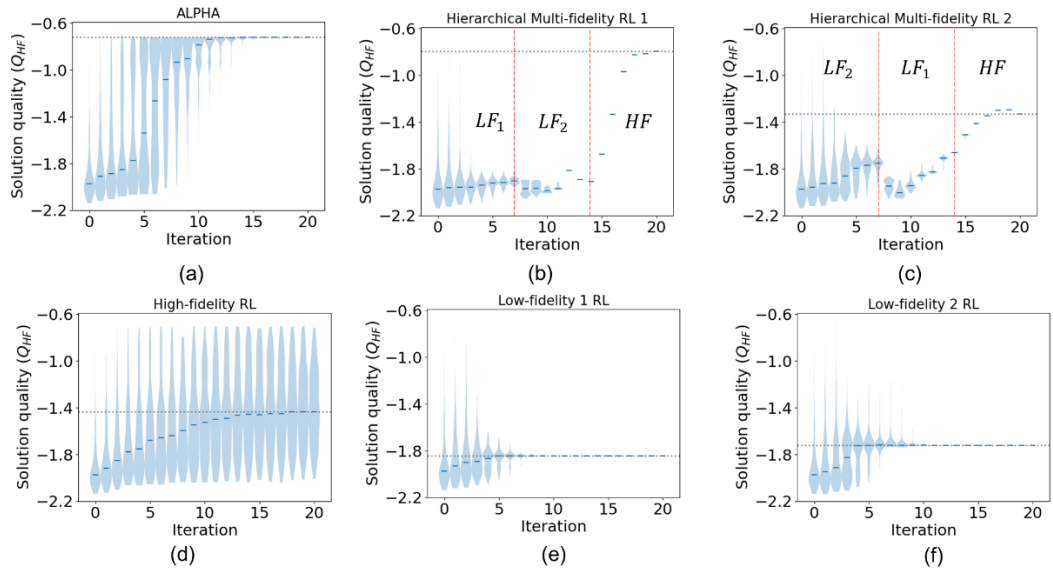

FIGURE 9: TRAINED POLICIES ARE EVALUATED BY PASSING SEEDS (VERTICAL LINE INDICATES FIDELITY SWITCH; VIOLIN PLOTS INDICATE MEDIANS)

The quality of the solutions for all the cases is better than the seed designs, indicated by the upward trend in all plots. However, the final solution qualities are significantly higher for the ALPHA case and Hierarchical MFRL 1 case as compared to all other cases. Further, the nature of the quality-iteration plots is drastically different between these two high performance cases. For the ALPHA case (Fig. 9(a)), there is a high dispersion in initial iterations followed by convergence to a high-quality solution. For the Hierarchical MFRL 1 case (Fig. 9(b)), the quality rises slowly when the low-fidelity models are operational. Further, a steep rise in quality is observed when the agent switches to the high-fidelity model. This difference between the agents is also reflected in the representative trajectories in Figure 10. Specifically, the adaptive agent follows a more direct path from the seed to the global optimum at A $(0.5, 0.5)$. This contrasts with the Hierarchical MFRL 1 case wherein agent first moves towards the point C $(-0.3, -0.3)$ as per the $LF_1$ model followed by a drastic change in direction as guided by the $LF_2$ model for the next few iterations. Further, it changes direction again as per the $HF$ model to converge near its global optimum at A (0.5,0.5). For the Hierarchical MFRL 2 case (Fig. 9(c)), the quality again rises slowly when the low-fidelity models are operational. Further, after switching to the high-fidelity model, a moderate increase in quality is observed. In Figure 10, this agent first moves towards the point D (0.3,0.3) as per the $LF_2$ model followed by a drastic change in direction as guided by the $LF_1$ model for the next few iterations. Further, it continues in that direction to converge to a nearby high-quality local optimum at B $(-0.5, -0.5)$ as guided by the $HF$ model. These trends of the hierarchical cases indicate that the performance of the underlying framework is sensitive to the ordering as well as the proportions of model usage. This contrasts with the adaptive framework which is not restricted by a predefined schedule of model usage. For the High-fidelity RL case (Fig. 9(d)), we observe a steady increase up to a moderate solution quality, albeit with high variance. For the illustrated representative trajectory, we observe convergence to a local optimum for this case. This is potentially because the agent has not fully learned the underlying complexities of the high-fidelity model within the 300 episodes. Lastly, for the cases that just utilize one of the low-fidelity models (Figs. 9(e) and 9(f)), the agents converge to low quality solutions based on the operational model. The trajectories corresponding to these cases converge to the optima of the low-fidelity models at approximately C $(-0.3, -0.3)$ and D (0.3,0.3), respectively.

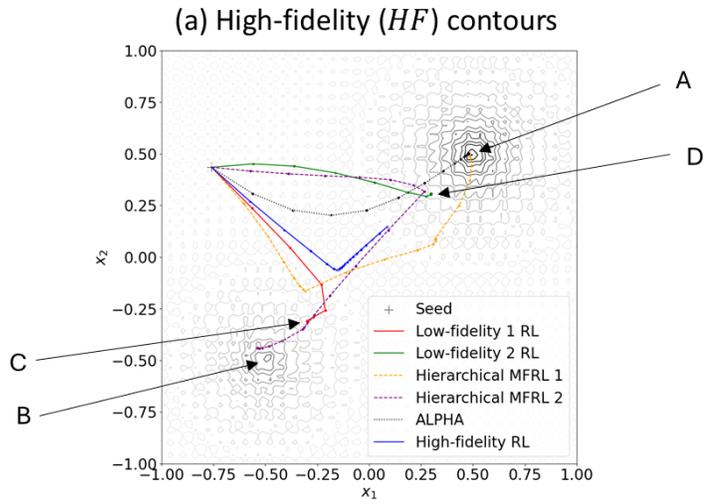

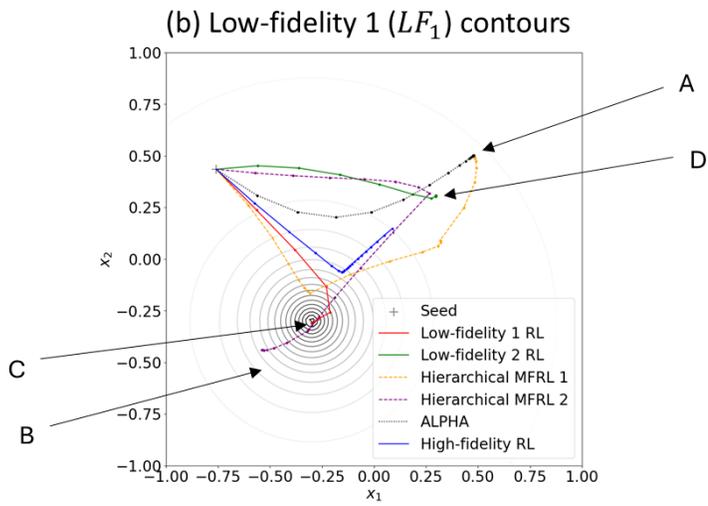

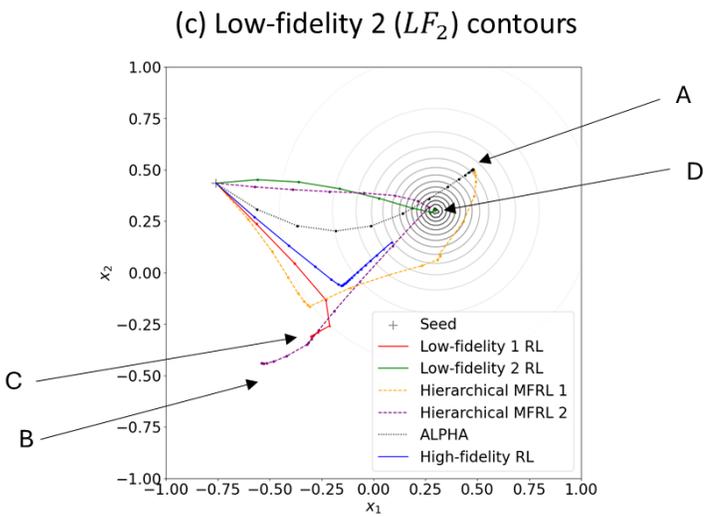

*FIGURE 10: REPRESENTAIVE EVALUATION TRAJECTORIES*

To understand the adaptive learning process of the ALPHA framework, the proportion of usage of models is evaluated across time. Figure 11 shows the evolution of model usage across the training of the agent with four distinct regimes $(R_1, R_2, R_3, R_4)$ in the trends. Figure 12 shows the policy maps corresponding to these regimes.

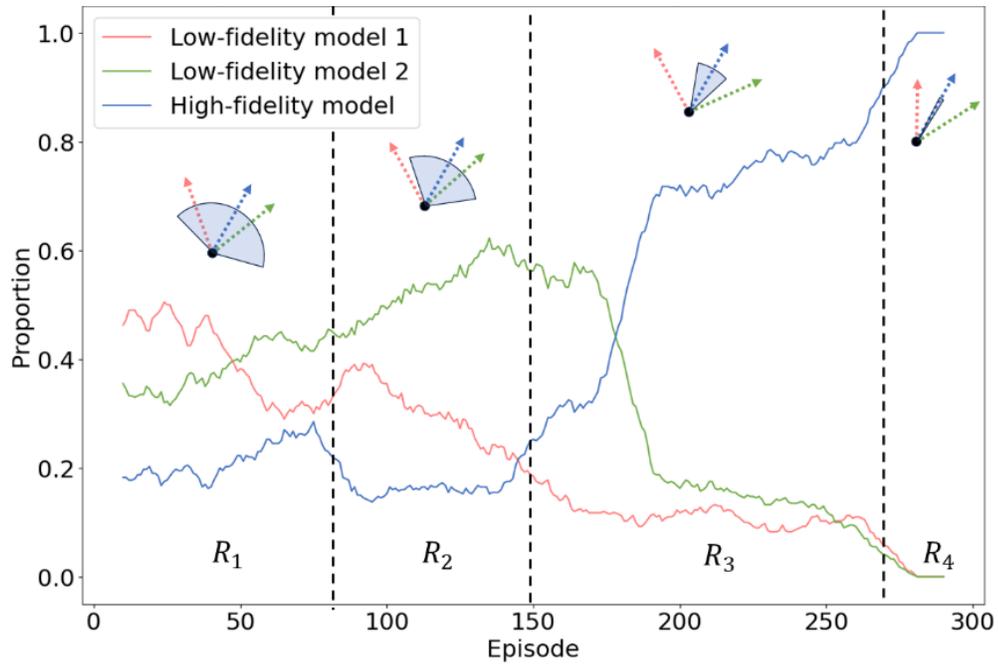

FIGURE 11: MODELS ARE ADAPTIVELY CHOSEN ACROSS TRAINING TIME

In the first regime $R_1$, the alignment threshold for the policies is the most relaxed. Further, the representative policy map at episode 40 in Figure 12 reveals that the agents have only learned some broad patterns, often leading to the $\epsilon$-greedy case where both $LF$ policies are aligned. Accordingly, the proportion of model usage reflects the corresponding probability values of model selection on average. In the next regime $R_2$, we observe an increase in the usage of the $LF_2$ model and a complementary decrease in the usage of the $LF_1$ model at tighter thresholds values. This is in correspondence with the policy map at episode 120 in Figure 12. Specifically, we observe that $LF_2$ policy is more aligned to the $HF$ policy than the $LF_1$ policy in the first, second and fourth quadrants of the problem domain. In the regime $R_3$, the usage proportion of both low-fidelity models drops with a complementary increase in the high-fidelity model usage. This is because of the misalignment of both $LF$ policies at even tighter threshold values, as depicted in the policy map at episode 220 in Figure 12. The last regime $R_4$ serves to increasingly refine the learning of the $HF$ policy with negligible influence of the $LF$ models. The final $HF$ policy as depicted in the policy map at 300 episodes converges to the global optimum at $(0.5, 0.5)$.

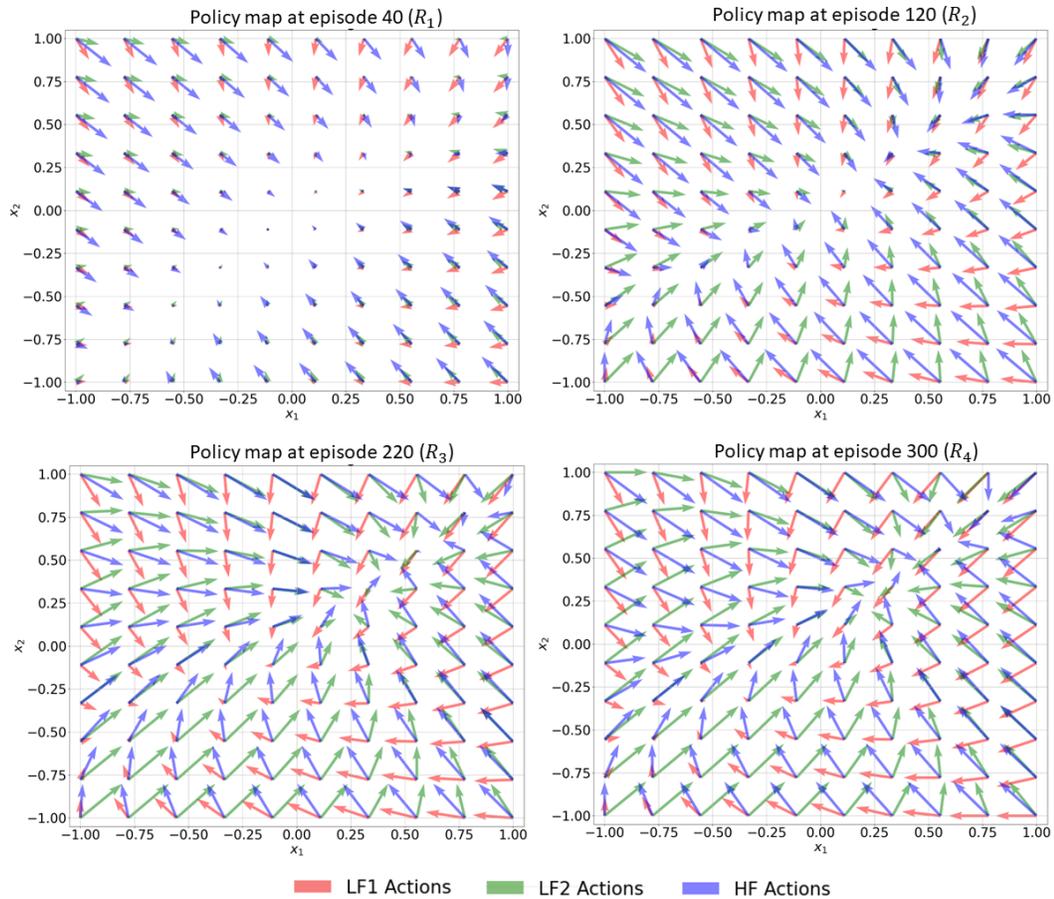

*FIGURE 12: EVOLUTION OF POLICIES ACROSS TRAINING*

To further understand the adaptive nature of model selection across search space, we employ Moran's I to analyze the spatial autocorrelation of model usage proportions within a grid space of the scaled problem domain. Figure 13 shows the model usage proportions in this grid space. For the $LF_1$ model grid, Moran's I is 0.4463 with a p-value of less than 0.0001, indicating a moderate and statistically significant spatial clustering of model usage. This corroborates the observation that the $LF_1$ model is most frequently used in the third quadrant. Further, this aligns with the policy maps in Figure 12 and can be attributed to the proximity of the $LF_1$ model optimum at $(-0.3, -0.3)$ to a high-quality local optimum of the $HF$ model at $(-0.5, -0.5)$. For the $LF_2$ model grid, Moran's I is 0.1161 with a p-value of 0.0739, suggesting weak and statistically insignificant clustering of model usage. This corroborates with the observation that the $LF_2$ model is widely used in first, second and fourth quadrants except in the proximity of $(0.5, 0.5)$. This aligns with the policy maps in Figure 12 and can be attributed to the proximity of the $LF_2$ model optimum at $(0.3, 0.3)$ to the global optimum of the $HF$ model at $(0.5, 0.5)$. For the $HF$ model, Moran's I is 0.3556 with a p-value of 0.0002, revealing moderate and statistically significant clustering of the HF model. This corroborates with the observation that the $HF$ model is dominantly used in the region surrounding this global optimum at $(0.5, 0.5)$. This can be attributed to the increased accuracy required near the global optimum, necessitating the high-fidelity model's precision. To summarize, low-fidelity models are used to explore broader regions, while the high-fidelity model focuses on critical areas, enhancing the efficiency and accuracy of the learning process. These trends showcase the ALPHA framework's adaptive use of multiple low-fidelity models alongside a high-fidelity model to achieve targeted learning across the search space.

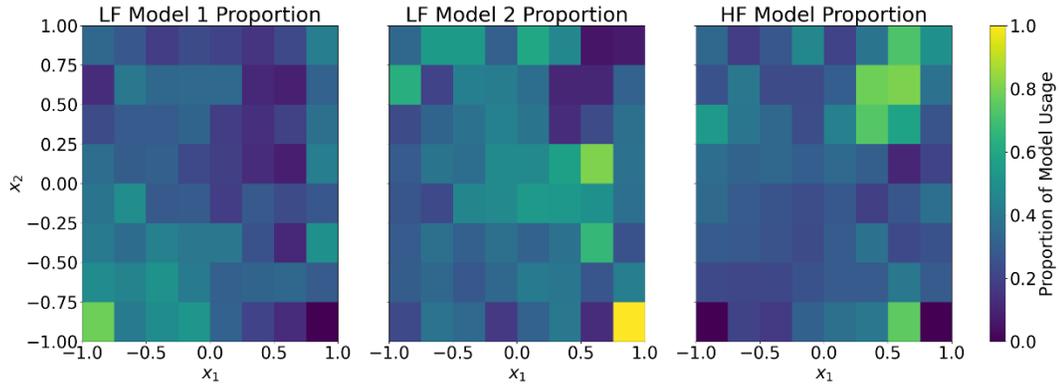

FIGURE 13: MODELS ARE ADAPTIVELY CHOSEN ACROSS SEARCH SPACE

To contextualize the quality-efficiency tradeoff of the ALPHA framework, the total time for evaluating the objective, and solution quality as measured by the high-fidelity model ($Q_{HF}$) was computed for all the agents. Figure 14 illustrates this tradeoff. Specifically, Fig. 14(a) shows the distribution of the quality of seeds and the solutions obtained using all the trained agents. Fig. 14(b) shows the total time required to evaluate the objective values during training. For the low-fidelity cases, we observe a low-quality of solutions obtained at a low computational cost. For the hierarchical cases, while both have a moderate computational expense, the solution quality is sensitive to the ordering of models. For the ALPHA case, we observe a high solution quality at a computational expense greater than the individual hierarchical cases. However, it is important to note that exploring different hierarchical schedules could incur additional time. Lastly, the high-fidelity case has the highest computational expense with a high variance in solution quality. This is potentially because the agent has not fully learned the underlying complexities of the high-fidelity model within the limited number of episodes. While training for longer may result in convergence to the global optimum, such extensive training will lead to even higher computational expense. These results showcase the capability of the proposed adaptive multi-fidelity framework to effectively balance solution quality with computational expense without the need for explicitly scheduling models.

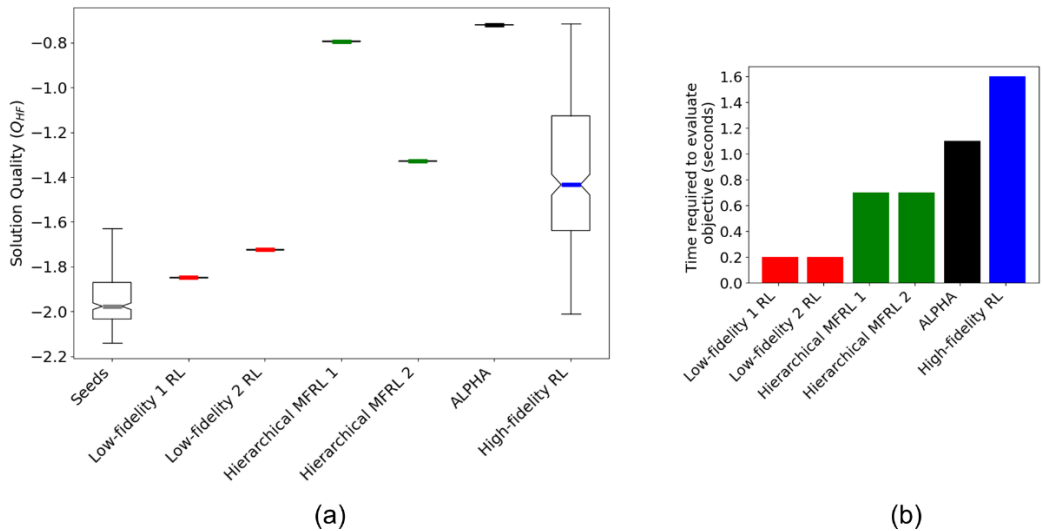

FIGURE 14: SOLUTION QUALITY-EFFICIENCY TRADEOFF

**5.2 Octocopter design problem**

The RL policies were trained and evaluated with the proposed adaptive multi-fidelity RL framework (ALPHA), a hierarchical multi-fidelity RL framework, and with each of the models individually, similar to the previous problem. However, 1200 seed points and episodes were utilized for training and evaluating all the policies. A higher number was used to account for the high-dimensionality and complexity of the design problem.

The results of the evaluation for the six cases (ALPHA, Hierarchical MFRL 1, Hierarchical MFRL 2, High-fidelity RL, Low-fidelity 1 RL, and Low-fidelity 2 RL) as per the high-fidelity model are shown in Figure 15. Furthermore, Figure 16 depicts representative evaluation trajectories for all cases in a reduced dimensional space of the scaled problem domain. Specifically, this reduced space was prepared using PCA on all the evaluation trajectories obtained from all the agents and the search data of the adaptive agent. The latter is integrated into the evaluation data to better understand the adaptive use of different models in the search space as discussed further in this section.

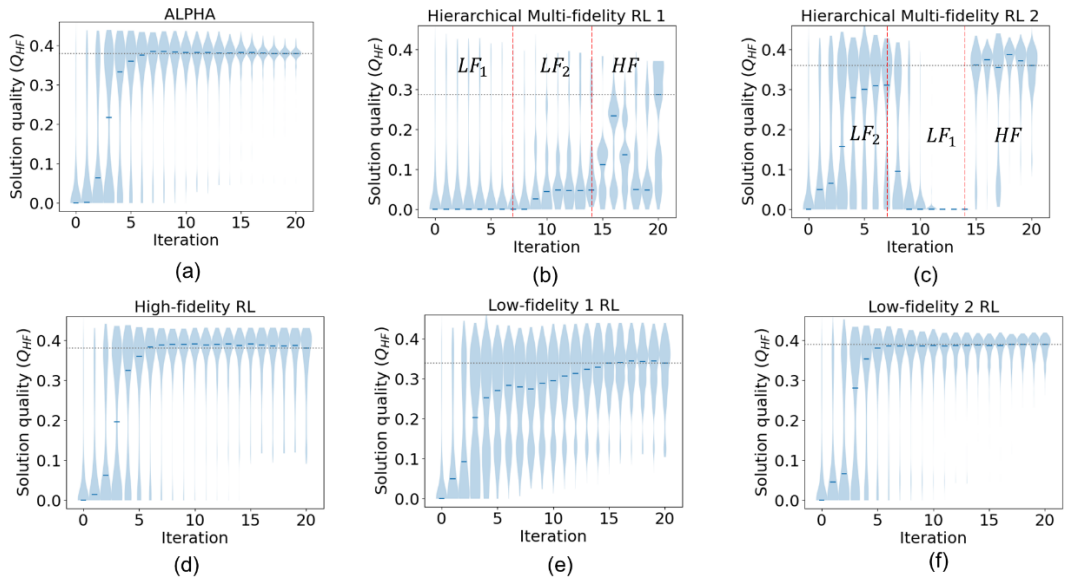

FIGURE 16: TRAINED POLICIES ARE EVALUATED BY PASSING SEEDS (FINAL SOLUTION QUALITY DISTRIBUTIONS SHOWN)

The quality of the solutions for all the cases is better than the seed designs, albeit with drastically different trends across the agents. For the ALPHA case (Fig. 16(a)), there is a high dispersion in initial iterations followed by convergence to a high quality with low variance. This indicates that the model can consistently find high-quality solutions. For the Hierarchical MFRL 1 case (Fig. 16(b)), the quality does not improve when the $LF_1$ model is operational. Further, the quality rises slowly when the $LF_2$ model is operational. Lastly, a steep rise to a moderate quality is observed when the agent switches to the high-fidelity model. However, this rise is not consistent across the iterations of the episode. For the Hierarchical MFRL 2 case (Fig. 15(c)), the quality rises significantly with a high dispersion in initial iterations when the $LF_2$ model is operational. Further, after switching to the $LF_1$ model, the quality drops to a poor value. Lastly, the quality rises to a high value with the use of the $HF$ model similar to the adaptive case (Fig. 15(a)), albeit with a higher variance. These trends of the hierarchical cases indicate that the performance of the underlying framework is sensitive to the ordering as well as the proportions of model usage. This contrasts with the adaptive framework which is not restricted by a predefined schedule of model usage. For the High-fidelity RL case (Fig. 16(d)), we observe a trend similar to the ALPHA case (Fig. 16(a)) with a high solution quality, albeit with a higher variance. This is potentially because the agent has not fully learned the underlying complexities of the high-fidelity model within the 1200 episodes. Lastly, for the cases that just utilize one of the low-fidelity models (Figs. 16(e) and 16(f)), the agents converge to high-quality solutions similar to the adaptive case (Fig. 16(a)). However, they both have higher variance than the adaptive case. In summary, the adaptive model stands out due to yielding both a high solution quality and minimal variability.

The difference between the agents is also reflected in the representative trajectories in Figure 17. Specifically, the ALPHA and Low-fidelity 2 RL agents follow paths that are most aligned to the high-fidelity agent. This corroborates with the high solution qualities in Figs. 16(a), 16(d), and 16(f). The Low-fidelity RL 1 agent converges in a different region. This is potentially a local optimum and is also reflected by the slightly lower solution quality value in Fig. 16(e). For the Hierarchical MFRL 1 case, while the overall nature of the trajectory is similar to the high-fidelity case, its orientation has significant deviation. This contrasts with the Hierarchical MFRL 2 case wherein agent first moves as per the $LF_1$ model followed by a drastic change in direction as guided by the $LF_2$ model for the next few iterations. Further, it changes direction again as per the $HF$ model to converge to a different region. This is also reflected in the solution quality plots in Figs. 16(b) and 16(c). These trends of the hierarchical cases reinforce that the performance of the underlying framework is sensitive to the ordering as well as the proportions of model usage. This contrasts with the adaptive framework which is not restricted by a predefined schedule of model usage.

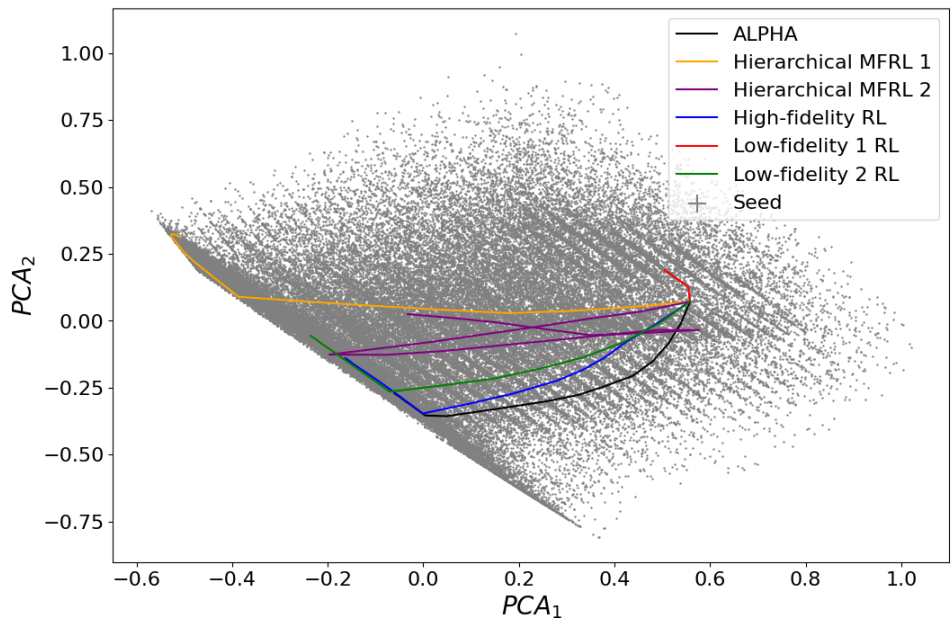

FIGURE 17: REPRESENTAIVE EVALUATION TRAJECTORIES

The proportion of usage of different models is also evaluated across time, similar to the previous case study. Figure 18 shows the evolution of model usage across the training of the agent with four distinct regimes ($R_1, R_2, R_3, R_4$) in the trends. Unlike the previous case study, we do not inspect the policy maps due to the high dimensionality of the problem.

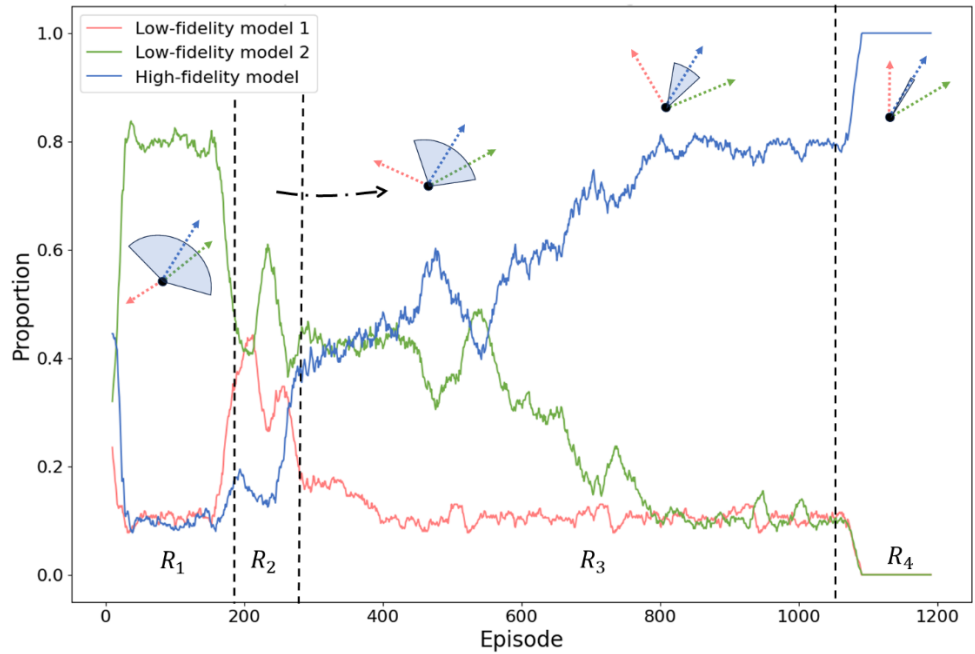

FIGURE 18: MODELS ARE ADAPTIVELY CHOSEN ACROSS TRAINING TIME

We observe that the model usage patterns between the three models are much more dynamic than the previous case study. Initially in the regime $R_1$, the $LF_2$ model is heavily utilized, with its proportion quickly rising above 0.8. Meanwhile, the high-fidelity model and low-fidelity model 1 are used sparingly. This is potentially because the $LF_2$ policy is more aligned to the $HF$ policy than the $LF_1$ policy, even with a loose initial alignment threshold. As training progresses into regime $R_2$, the usage of both the low-fidelity models begins to fluctuate, essentially complementing each other for a few episodes. This is indicative of the varied alignment of the low-fidelity policies with the high-fidelity policy. In the regime $R_3$, there is a decrease in the usage of both low-fidelity models, with the decrease in the $LF_1$ model much more rapid the $LF_2$ model. This decrease coincides with an increase in the proportion of the $HF$ model. This transition suggests that as the alignment threshold tightens the algorithm relies on more accurate feedback from the $HF$ model. Moreover, the different rates of decrease of the $LF$ models correspond to the varied alignment of the $LF$ policies with respect to the $HF$ policy. The last regime $R_4$ serves to increasingly refine the learning of the HF policy with negligible influence of the $LF$ models.

To further understand the adaptive nature of model selection across search space, we employ Moran's I to analyze the spatial autocorrelation of model usage proportions within a grid space spanning the PCA space in Figure 17. Figure 19 shows the model usage proportions in this grid space. For the $LF_1$ model grid, Moran's I is 0.0558 with a p-value of 0.0184, indicating negligible spatial clustering of model usage. This corroborates with the observation that the $LF_1$ model is sparingly used in most cells except some cells at the boundaries of the explored space. In contrast, the grids of the $LF_2$ and $HF$ models exhibit higher Moran's I values of 0.1678 and 0.2004, respectively, both with p-values of less than 0.0001, reflecting weak yet statistically significant clustering. Further, we observe that the $LF_2$ model is most frequently used in the first, second and fourth quadrants. Lastly, the $HF$ model is most frequently used in the third quadrant and is also utilized to a reasonable extent in the second quadrant. The exclusive high usage in the third quadrant can potentially be attributed to the increased accuracy required to refine the policy, necessitating the high-fidelity model's precision. While a deeper understanding of the underlying patterns is challenging due to the high-dimensionality and complexity of the problem, these trends showcase the ALPHA framework's adaptive use of multiple low-fidelity models alongside a high-fidelity model to achieve targeted learning across the search space.

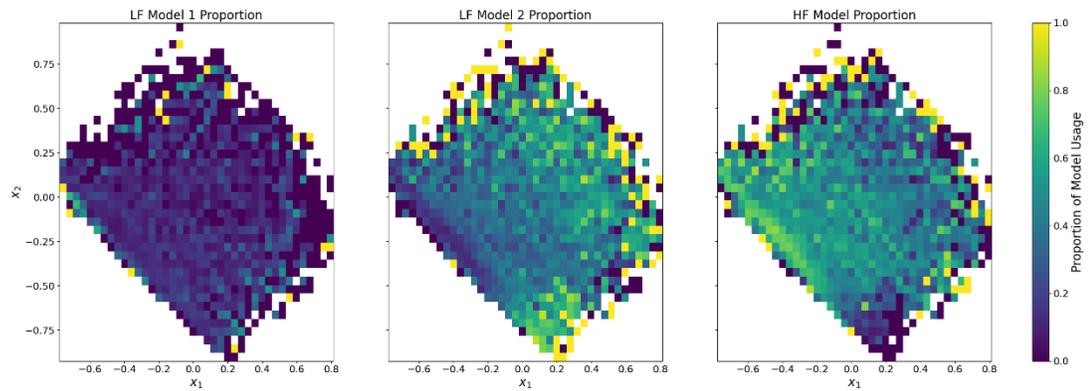

FIGURE 19: MODELS ARE ADAPTIVELY CHOSEN ACROSS SEARCH SPACE

To contextualize the quality-efficiency tradeoff of the ALPHA framework, the total time for evaluating the objective, and solution quality as measured by the high-fidelity model ($Q_{HF}$) was computed for all the agents. Figure 20 illustrates this tradeoff. Specifically, Fig. 20(a) shows the distribution of the quality of the solutions obtained using all the trained agents. Fig. 20(b) shows the total time required to evaluate the objective values during training. First, we observe that all the agents find high performance solutions on average. However, there is significant variation in the distribution of solution quality. For the Low-fidelity 1 RL, Hierarchical MFRL 1, Hierarchical MFRL 2 and High-fidelity RL cases, the variance is significantly higher than the other two cases. Amongst these two cases, the ALPHA case has a lower variance than the Low-fidelity 2 RL case. This shows that the adaptive agent can uniquely find consistent high performance solutions as opposed to other agents. In context of computational efficiency, the costs of the cases that just use the $LF$ models are negligible. For the hierarchical cases, while both have a moderate computational coat, the solution quality is sensitive to the ordering of models. For the ALPHA case, we observe a high solution quality at a computational expense greater than the individual hierarchical cases. However, it is important to note that exploring different hierarchical schedules could incur additional time. Lastly, the High-fidelity RL case has the highest computational expense. Notably, the high variance in quality of the high-fidelity agent is potentially because the agent has not fully learned the underlying complexities of the high-fidelity model within the limited number of episodes. While training for longer may result in convergence (similar to the ALPHA agent), such extensive training will lead to even higher computational expense. These results showcase the capability of the proposed adaptive multi-fidelity framework to effectively balance solution quality with computational expense without the need for explicitly scheduling models.

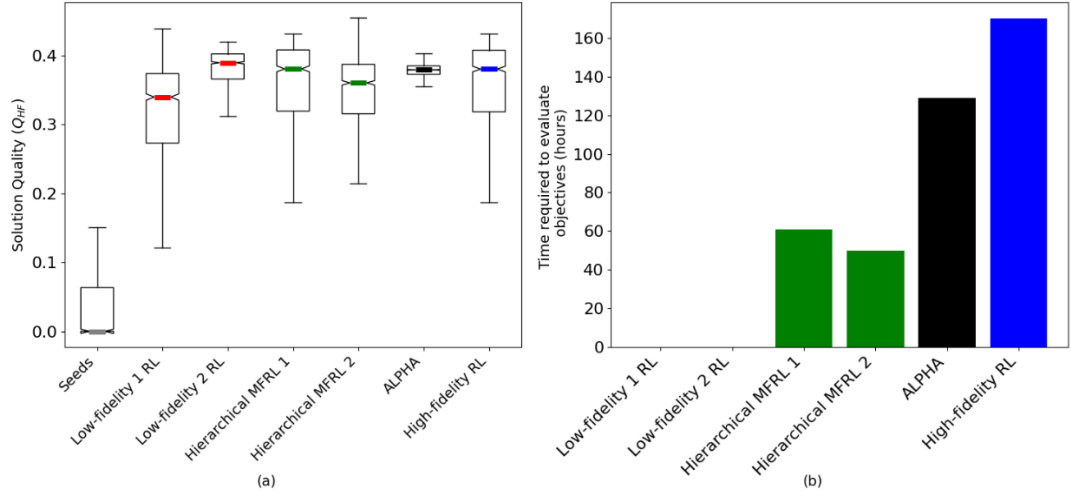

FIGURE 20: SOLUTION QUALITY-EFFICIENCY TRADEOFF

## 6. CONCLUSION

This research presents a novel multi-fidelity reinforcement learning framework, ALPHA (Adaptively Learned Policy with Heterogeneous Analyses), that efficiently learns a high-fidelity policy by adaptively leveraging heterogeneous low-fidelity models alongside a high-fidelity model. We showcase the effectiveness of the framework in two case studies involving analytical test optimization and octocopter design problems. In contrast to traditional hierarchical multi-fidelity methods, our approach adapts dynamically, prioritizing different models based on the alignment of low-fidelity policies with the high-fidelity policy. This adaptive use eliminates the need to schedule models and allows for targeted learning in critical areas while reducing computational costs in less critical regions. Moreover, incorporating aligned low-fidelity experience data in the learning of the high-fidelity agent leads to a unified high-fidelity policy capable of comprehensively exploring the design space. The results demonstrate that the unified high-fidelity policy consistently yields high quality solutions at a reduced computational expense. Furthermore, the adaptive agents find more direct paths to high-performance solutions, showing superior convergence behavior compared to hierarchical agents.

Future work should explore enhancements to improve the adaptivity and the resultant solution quality-efficiency tradeoff of the proposed multi-fidelity framework. For instance, there is potential to utilize cost-weighted probabilities in model selection when multiple low-fidelity policies align with the high-fidelity policy. Moreover, different alignment metrics and threshold scheduling techniques could improve the adaptivity of the approach. While this research demonstrates the framework's ability to handle an arbitrary number of heterogeneous models, future studies could investigate leveraging hierarchical structures within some of these models. By directly managing the hierarchy while still addressing model heterogeneity, the framework's performance could be significantly enhanced. This work also assumes that each simulator can operate on a common design representation. Future work should explore extensions to the framework to be interoperable across representations.

Comparative studies with other RL-based methods that explicitly model errors across different fidelity levels would provide valuable insights into the relative strengths and limitations of the proposed approach. Additionally, investigating the applicability of various prevailing multi-fidelity model management strategies [8,9] beyond those in the realm of RL could further enhance the framework. Such studies could explore the trade-offs between computational efficiency and solution quality across different methodologies, providing a comprehensive comparison with the proposed adaptive approach. Adapting the framework to multi-disciplinary optimization scenarios [53] could exploit the interactions between various disciplines, leading to more effective search strategies. Furthermore, incorporating budget-aware extensions to optimize resource allocation would significantly enhance the practical utility of the framework, particularly in resource-constrained scenarios. These advancements can ultimately lead to more efficient and adaptive agents capable of tackling complex engineering design challenges across various domains.

**ACKNOWLEDGMENT**

The authors are grateful to Srikanth Devanathan of Dassault Systèmes for his feedback on early versions of this work.

**FUNDING**

This material is based upon work supported by the Defense Advanced Research Projects Agency through cooperative agreement FA8750-20-C-0002. Any opinions, findings, and conclusions or recommendations expressed in this work are those of the authors and do not necessarily reflect the views of the sponsors.